\pdfoutput=1
\documentclass[11pt]{article}

\usepackage{emnlp2021}

\usepackage{times}
\usepackage{latexsym}

\usepackage[T1]{fontenc}
\usepackage[utf8]{inputenc}

\usepackage{microtype}

\usepackage{graphicx} % DO NOT CHANGE THIS
\usepackage{amsmath}
\usepackage{verbatim}
\usepackage{multirow}

\usepackage{cleveref}
\crefname{section}{§}{§§}
\Crefname{section}{§}{§§}

\def \modelName {WebSRC}

\title{\modelName: A Dataset for Web-Based Structural Reading Comprehension}

\author{Xingyu Chen, Zihan Zhao, Lu Chen\thanks{\ \ The corresponding authors are Lu Chen and Kai Yu.}, \\ {\bf Jiabao JI, Danyang Zhang, Ao Luo, Yuxuan Xiong and Kai Yu\footnotemark[1]}\\
  X-LANCE Lab, Department of Computer Science and Engineering\\
  MoE Key Lab of Artificial Intelligence, AI Institute, Shanghai Jiao Tong University\\
  Shanghai Jiao Tong University, Shanghai, China\\
  State Key Lab of Media Convergence Production Technology and Systems, Beijing, China\\
  \{galaxychen, zhao\_mengxin, chenlusz\}@sjtu.edu.cn, \\ \{sjcs\_jijiabao, zhang-dy20, wenzelmetternich, xiongyx, kai.yu\}@sjtu.edu.cn}

\begin{document}
\maketitle
\begin{abstract}
 Web search is an essential way for humans to obtain information, but it's still a great challenge for machines to understand the contents of web pages. In this paper, we introduce the task of structural reading comprehension (SRC) on web. Given a web page and a question about it, the task is to find the answer from the web page. This task requires a system not only to understand the semantics of texts but also the structure of the web page. Moreover, we proposed \modelName, a novel \textbf{Web}-based \textbf{S}tructural \textbf{R}eading \textbf{C}omprehension dataset.
\modelName{} consists of 400K question-answer pairs, which are collected from 6.4K web pages. Along with the QA pairs, corresponding HTML source code, screenshots, and metadata are also provided in our dataset. Each question in \modelName{} requires a certain structural understanding of a web page to answer, and the answer is either a text span on the web page or yes/no.
We evaluate various baselines on our dataset to show the difficulty of our task. We also investigate the usefulness of structural information and visual features. Our dataset and baselines have been publicly available\footnote{https://x-lance.github.io/WebSRC/}.
\end{abstract}

\section{Introduction}
\label{sec:intro}

\begin{figure}[h]
\centering
\includegraphics[width=1\columnwidth]{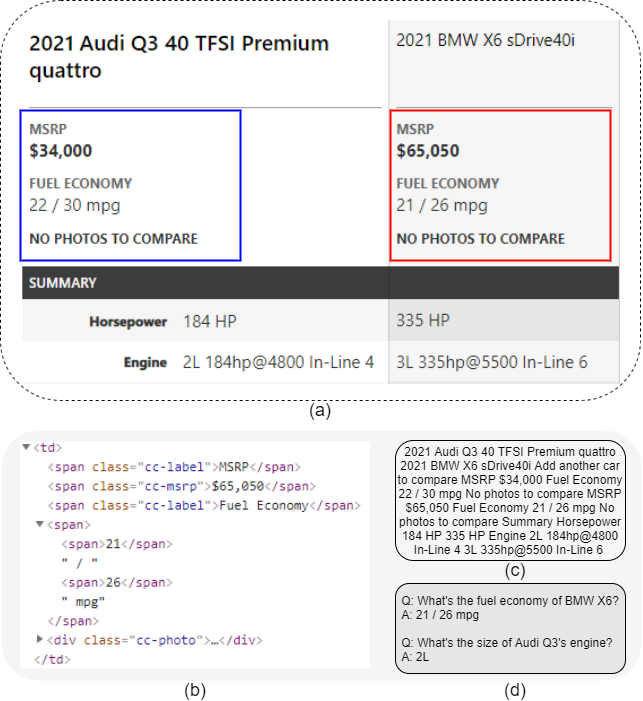} % Reduce the figure size so that it is slightly narrower than the column. Don't use precise values for figure width.This setup will avoid overfull boxes.
\caption{Examples for a web page. (a) is the original web page. (b) is the HTML code for the content in the red box. Each HTML tag begins with a starting tag and ends with a closing tag (with a slash in tag). \textit{<td>} stands for a table cell and \textit{<span>} stands for a content span. (c) shows the text extracted from the web page. (d) contains some sample questions for the web page.}
\label{fig-example}
\end{figure}

\begin{table*}[h]
    \begin{center}
    \begin{tabular}{ |c||c|c|c|c|c| } 
     \hline
     Datasets & \#domain & \#website & Task & \#Query & With Image \\
     \hline
     WEIR\citep{bronzi2013extraction} & 4 & 40 & ClosedIE & 32 & No\\ 
     SWDE\citep{hao2011one} & 8 & 80 & ClosedIE & 32 & No\\ 
     Expanded SWDE\citep{lockard2019openceres} & 3 & 21 & OpenIE & 748 & No\\ 
     \modelName{}(Ours) & 11 & 70 & QA & 2735 & Yes\\ 
     \hline
    \end{tabular}
    \caption{The comparison with datasets with HTML. The query in ClosedIE is the attributes needed to be extracted, in OpenIE is predicates, and in WebSRC is the questions before data augmentation. }
    \label{comparision}
    \end{center}
\end{table*}

Web pages are the most common source of human knowledge and daily information. With the help of modern search engines, people can easily locate web pages and find information by simply typing some keywords. However, traditional search engines only retrieve web pages related to the query and highlight the possible answers \citep{chen2018neural}, they can't understand the web pages and answer the query based on contents. The rapid development of question answering systems and knowledge graphs enables search engines to answer simple questions directly \citep{chakraborty2019introduction}, but they still fail to perform question answering on arbitrary web pages. The difficulty lies in the variety of web pages and the complexity of the web layouts, which requires a system not only to consider the text but also the structures of web pages.

There are two kinds of structures for each web page: \textit{spatial} structure and \textit{logical} structure. The spatial structure is how the information is visually organized, and the logical structure is how the information is organized by semantics.
Figure \ref{fig-example}(a) shows the spatial structure of the web page, e.g., how the texts are arranged and what are their relative positions. The logical structure can be deduced by the spatial structure and the semantics of the texts. For example, this image introduces the information about two cars, with car names at the top followed by the detailed specifications.
A human can easily answer the questions in Figure \ref{fig-example}(d) by referring to the relevant section in the logical structure. But for computers, it's hard to understand the logical structure by just taking the spatial structure (the image) as input due to the lack of common sense. Computers need to infer the answers from the font size, the color and the spatial relations between texts, let alone they need to extract texts from the image and understand them.

An alternative way is to utilize the text from web page. Figure \ref{fig-example}(c) shows the texts extracted from Figure \ref{fig-example}(a). As we can see, the layout structure is lost in the plain text, and the text is just a concatenation of short phrases without a meaningful context. It would be difficult to answer questions only based on such texts. 
Besides texts, we can also parse the HTML (Hypertext Markup Language) document, i.e. the source code of the web page. It describes the structure of the webs page and uses \textit{HTML elements (tags)} to display the contents. We will use the term \textit{tag} and \textit{element} in this paper interchangeably. Figure \ref{fig-example}(b) shows the HTML code corresponding to the part of the web page highlighted in the red box. 
HTML is a kind of semi-structured document \citep{buneman1997semistructured}, where tags with different structural semantics serve as separators. It's also called the ``self-describing'' structure. 
An HTML document can be parsed into a tree-like structure called DOM\footnote{https://en.wikipedia.org/wiki/Document\_Object\_Model} (Document Object Model), where the tree nodes are elements in the HTML, and texts are all leaf nodes in the tree. An HTML DOM tree can serve as a structural representation of the web page, where visually similar items on the web page would be sub-trees with similar structures. For example in Figure \ref{fig-example}, the HTML structure is identical for the segment in the blue and red boxes. They are only different in the text.
However, due to the complexity of rendering HTML code into a web page, a single HTML would not be enough to represent the full logic structure of the web page. For example, in Figure \ref{fig-example}(b), the four \textit{<span>} are in the same spatial level of the DOM tree, but they play different semantic roles in the web page, i.e. the first span indicates an attribute and the second contains the corresponding value. We need to leverage both the visual and structural information to gain a comprehensive understanding.

To promote researches in question answering on web pages, we introduce \modelName, a dataset for reading comprehension on structural web pages. The task is to answer questions about web pages, which requires a system to have a comprehensive understanding of the spatial structure and logical structure. \modelName{} consists of 6.4K web pages and 400K question-answer pairs about web pages. For each web page, we manually chose one segment from it and saved the corresponding HTML code, screenshot, and metadata like positions and sizes.
Questions in \modelName{} were created for each segment. Answers are either text spans from web pages or yes/no. Taking the HTML code, screenshot, metadata as well as question as input, a model is to predict the answer from the web page. The comparison of \modelName{} with other datasets with HTML documents is illustrated in Table \ref{comparision}. Our dataset is the only one that provides HTML documents and images, and is larger in the number of domains and queries.

To summarize, our contributions are as follows: 
\begin{itemize}
    \item We proposed the task of structural reading comprehension (SRC) on web, which is a multi-modal machine reading comprehension task that focuses on understanding texts and screenshots on web pages.
    \item We created a large dataset for web-based structural reading comprehension consisting of 400K QAs and 6.4K web page segments, where HTML code and additional visual features are also provided.
    \item We evaluated several baselines on \modelName{} and the results showed that \modelName{} is highly different from the existing textual QA datasets and is challenging even for the leading pre-trained language model.
\end{itemize}

\section{Related Work}
Machine reading comprehension (MRC) models have achieved excellent performance on plain text corpus \citep{zeng2020survey} in recent years. Traditional datasets for machine reading comprehension \citep{talmor2019commonsenseqa,yang2018hotpotqa,rajpurkar2016squad,rajpurkar2018know,choi2018quac,reddy2019coqa,lai2017race} contain plain text passages and QAs about them. However, HTML code in the form of semi-structured documents is different from the ordinary textual corpus. Recently, multi-modal MRC has gained the interest of researchers. Multi-modal MRC datasets with both images and texts are proposed, such as MovieQA \citep{tapaswi2016movieqa}, TQA \citep{kembhavi2017you}, COMICS \citep{iyyer2017amazing} and RecipeQA \citep{yagcioglu2018recipeqa}. Images in these datasets provide different information from texts, and texts are supplementary descriptions for images. 
Text VQA \citep{mishraICDAR19,singh2019towards,docvqa_wacv} is a kind of VQA (visual question answering) task \citep{VQA}, whose task is to answer questions about a real-world image, and questions in this task are about the texts in the image. However, there is no existing text or layout description available in the image, but we can access them easily on web pages.

Information extraction for web pages has been investigated intensively \citep{chang2006survey}. Previous studies mainly focus on building templates for HTML DOM tree, called Wrapper Induction \citep{kushmerick2000wrapper, flesca2004web, kushmerick1997wrapper, muslea1999hierarchical}, or using well designed visual features like font sizes, element sizes, and positions \citep{zhu20052d, zhu2006simultaneous}. These methods require abundant human labor to label templates and analyze features, which makes it hard to generalize to unseen websites. \citet{10.1145/3453483.3454047} proposed a program synthesis based technique to extract web information. Some studies focused on recognizing tables from web pages \citep{zanibbi2004survey} and tried to model the physical and logical structure of tables in HTML, \citet{zhang-etal-2020-graph} proposed to use a graph to represent the table structure. 
Some web QA datasets are proposed \citep{dunn2017searchqa,joshi-etal-2017-triviaqa,dhingra2017quasar,li2016dataset}, but they only contain text snippets extracted from web pages. \citet{bronzi2013extraction} proposed a dataset called WEIR, consisting of 40 websites from 4 domains. \citet{hao2011one} proposed SWDE, which contains 124,291 web pages from 80 websites, and \citet{lockard2019openceres} expanded SWDE for openIE. All these datasets only contain HTML code for extraction, and the task is to extract pre-defined attributes of entities in web pages, e.g. the author of a book.
Layout analysis \citep{binmakhashen2019document} is the task to analyze document images like contracts, bills, and business emails. IIT-CDIP~\cite{DavidDLewis2006SIGIR_IIT_CDIP} and RVL-CDIP~\cite{AdamWHarly2015ICDAR_RVL_CDIP} are two datasets collected for document classification. \citet{GuillaumeJaume2019ICDARW_FUNSD} proposed FUNSD for form understanding and \citet{ZhengHuang2019ICDAR_SROIE} organized SROIE competition for receipt understanding.
PubLayNet~\cite{XuZhong2019ICDAR_PubLayNet} and DocBank~\cite{li2020docbank} are proposed to benchmark the task of layout recognition in academic papers. However, compared to the images in the layout analysis task, web pages are much more complex in organizing information. The terms to be recognized are relatively stable in layout analysis, while web pages may contain various information that is hard to be pre-defined.

\begin{figure*}
\centering
\includegraphics[width=1.8\columnwidth]{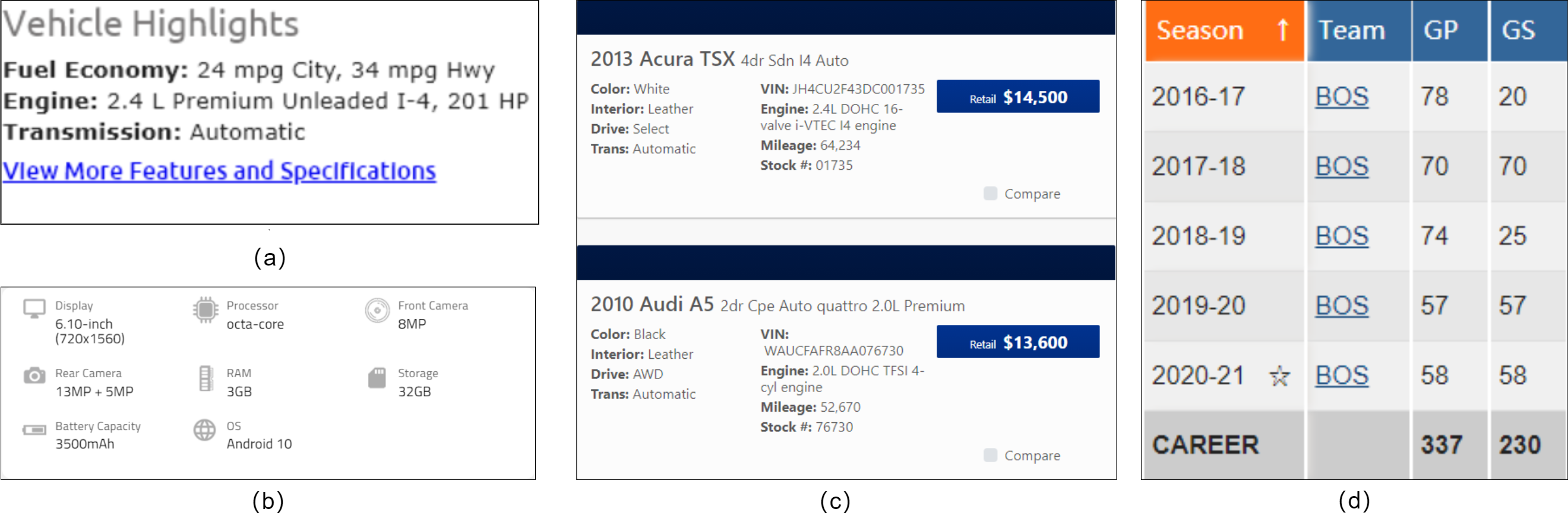} % Reduce the figure size so that it is slightly narrower than the column. Don't use precise values for figure width.This setup will avoid overfull boxes.
\caption{Examples for three types of web pages. (a) and (b) are web pages of type \textit{KV}, (c) is a web page of type \textit{comparison}, (d) is a web page of type \textit{table}.}
\label{website-example}
\end{figure*}

\section{Data Collection}
The construction of our dataset consists of five stages: \cref{web-page-selecting} web page selection, \cref{web-page-collecting} web page collection, \cref{question-labeling} question labeling, \cref{data-augmentation} data augmentation and \cref{final-review} final review. We will describe each stage in detail below.

\subsection{Task Definition}
The task of structural machine reading comprehension on web can be described as given the context $\mathcal{C}$ and a question $q$, predict the answer $a$. In our task, the context can be HTML code, screenshots, and the corresponding metadata. Denote the machine reading comprehension model as $\mathcal{F}$, our task can be formulated as:
%$$\mathcal{F}(\mathcal{C},q) = a$$
\begin{equation}
  \mathcal{F}(\mathcal{C},q) = a
\label{eq:task}
\end{equation}

\subsection{Locating text in HTML}
To precisely locate the text in HTML, we first define \textbf{the text of an HTML node}: the text of an HTML node is the concatenation of texts in its descendant nodes in the DOM tree, where the order of texts is derived by the depth-first search. With this definition, a text can be located by the tag containing the text, and the beginning position in \textit{the text of the node}.

\subsection{Web page selection}
\label{web-page-selecting}
In this phase, we choose websites for further data collecting. We are interested in the structure of the web page, so in the web page selection phase, we only focused on websites with a relatively complex structure and that have abundant information for question answering. We didn't choose websites with long textual paragraphs like Wikipedia, where the structure has little influence on understanding the content. We started from the website list of the SWDE \citep{hao2011one} dataset, which contains 80 websites from 8 domains. Websites on the list that are no longer available are dropped. We also expanded our website list by searching the domain keywords and selected the most relevant websites. In total, we obtained 70 websites from 11 domains.

We didn't use the whole web page but only chose some segments to build our dataset, because a complete web page may contain ads or additional structures like navigation tabs, which brings too much noise into the web page and makes the task much harder. We admit that in the real-world scenario we have to deal with the full web page, but we consider the problem of question-answering in full web pages can be modeled as a two-stage process: first, find the relevant segment in the web page and then answer the question based on the segment.
In this work, we will focus on learning the structure of a given web page segment and leave the segment locating problem as future work.

The choice of the segment is based on the type of web page. We category web pages into three types, \textit{KV}, \textit{comparison}, and \textit{table}, according to the different ways to display information. We will discuss different types of websites in detail below.

\textbf{KV} Information in this type of web page is presented in the form of ``\textit{key: value}'', where the \textit{key} is an attribute name and the \textit{value} is the corresponding value. See Figure \ref{website-example}(a) and Figure  \ref{website-example}(b) for illustration. This kind of web page can be found from the detail page of an entity, e.g. a car or a book. 
We choose the section that describes attributes about the entity from the web page.

\textbf{Comparison} This type is similar to type KV but with a major difference: web pages of type comparison contain several entities with the same attributes. For instance, in Figure \ref{website-example}(c), there are two cars with same attributes in the image and they form a comparison. We chose the segment that at least contains a comparison between two objects.

\textbf{Table} Web pages of this type use a table to present information. A table contains the comparison between rows naturally but unlike the type comparison, it uses a unified header to represent attributes and each row in the table only contains values. Figure \ref{website-example}(d) shows the statistics table of a basketball player. 
The segment we chose is the table area on the web page.

\begin{figure*}[h]
\centering
\includegraphics[width=2\columnwidth]{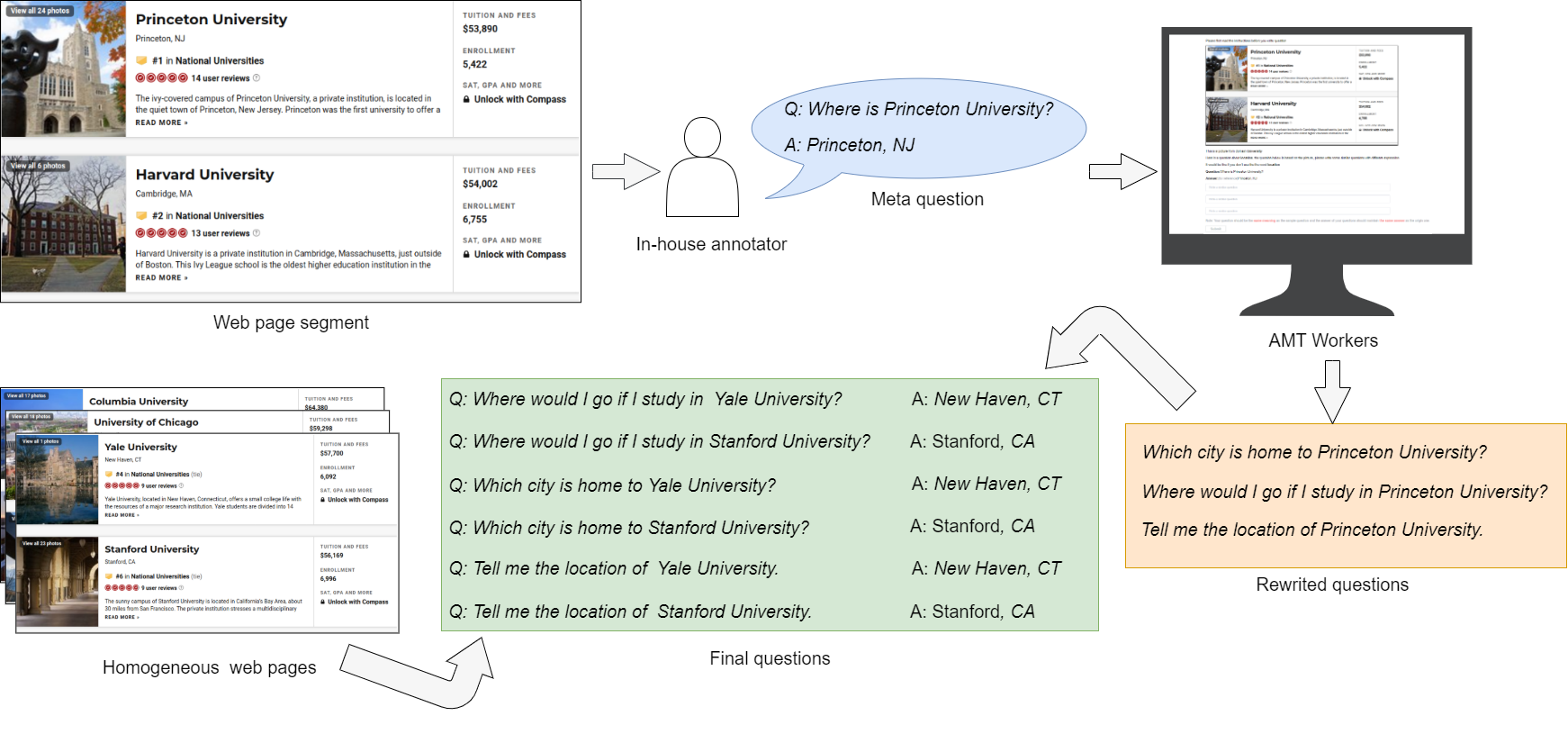} % Reduce the figure size so that it is slightly narrower than the column. Don't use precise values for figure width.This setup will avoid overfull boxes.
\caption{The pipeline for labeling questions and augmenting data.}
\label{fig-augmentation}
\end{figure*}

\subsection{Web page collection}
\label{web-page-collecting}
We recruited six computer science students with web crawling experience to collect the web pages. We first rendered the website in the headless Chrome browser, then for each segment, we manually wrote extracting code to crawl it. We saved the corresponding HTML and the screenshot of the segment, as well as additional metadata (including the location and size of each tag, the color and font of texts). We used Selenium\footnote{https://www.seleniumhq.org/} to collect all the data.

For segments of type comparison or type table, we would drop some objects in comparison or delete some rows in the table if the size of the segment is too large. 
We crawled homogeneous segments from 100 web pages under each website, each of them shares a common HTML structure but with different content. We obtained 6447 web pages after dropping some invalid web pages.
We removed all non-ascii characters and extra spaces in HTML. Tags that have little influence on HTML structure are removed, including the \textit{<script>} and the \textit{<style>}. Properties of HTML tags are also removed except for \textit{class}, \textit{id}, \textit{title} and \textit{aria-label}, for these properties often serve as descriptions of a tag. We added an additional attribute called \textit{tid} to all tags, which was used in locating tags with answers.

\subsection{Question Labeling}
\label{question-labeling}
We recruited three annotators to label questions and answers for each crawled segment. We showed screenshots to annotators and asked them to create questions about the content on the image. All questions should be answerable by the screenshot, and the answer should be a text shown in the image or yes/no. We asked annotators to create questions in the following style:
\begin{itemize}
    \item Ask questions about certain key-value pair. For example in Figure \ref{website-example} (a), \textit{what's the engine specification of this car?}
    
    \item Ask questions about certain object in the comparison. For example in Figure \ref{website-example} (c), \textit{what's the price of Audi A5?}
    
    \item Ask questions about a cell value in the table. For the table example in Figure \ref{website-example} (d), \textit{what's the GP score in 2017-18?}
    
    \item Ask questions with condition. For example in Figure \ref{website-example} (c), \textit{what's the price of the white car?} with a condition "white".
    
    \item Ask yes/no questions for confusing terms. For example in Figure \ref{website-example} (b), \textit{Is the storage 32GB?} asks about the storage size which is similar to the RAM.
\end{itemize}

We also asked annotators to label the answer in the HTML, including the answer text, the tag containing the answer (represented using \textit{tid}) and the beginning position of the answer in \textit{the text of the answer tag}. When creating questions, we also encourage annotators to ask questions that are meaningful from an actual end user's perspective.

We asked a different annotator to check if the question is followed one of the styles above and if the answer is a valid text in the segment or yes/no. We collected 460 unique questions for all segments, and we called these questions \textit{meta-question}s.

To enhance the diversity of question expression, we published a question rewriting task on Amazon Mechanical Turk (AMT) to polish meta-questions. Workers on AMT were shown a screenshot and a meta-question with the answer, and their task is to rewrite the given question without changing the meaning. We encouraged the worker to use more complex expressions and use synonyms for attributes if possible. Each worker should create three different versions of meta-questions. 191 workers participated in the rewriting task, and we asked another four annotators to filter questions with obvious grammar errors and inconsistent meaning. About $10\%$ questions are dropped after review. Examples of rewritten questions are shown in Figure \ref{fig-augmentation}. As we can see, annotators may change the way of asking, introduce subjunctive mood or change the question to an imperative sentence. We collected 2735 questions at this stage.

\subsection{Data Augmentation}
\label{data-augmentation}
Although the structure of different websites varies a lot, web pages under the same website have a similar structure. In this phase, we automatically applied collected questions to all homogeneous web pages. For each question, we manually created extracting rules to identify answers on different web pages. We generate new QA pairs for all web pages by replacing the original answer with the answer extracted from the homogeneous web page. If a question contains a specific entity name, e.g. a car name in the comparison, we also replace it with the actual entity on the corresponding web page.

After data augmentation, we obtained 400498 question-answer pairs in total. The whole process of question labeling and data augmentation is illustrated in Figure \ref{fig-augmentation}.

\subsection{Final review}
\label{final-review}
We wrote tests for the dataset to check the correctness of the label, the completeness of saved files and the format of dataset. We also sample 100 QA pairs from each website and asked four experienced annotators to double-check the correctness of semantics, e.g. whether the answer matches the question, whether the question is suitable for the web page. Cases with errors would send back to annotators for a new round of labeling.

\section{Dataset Analysis}
In this section, we conduct throughout analysis of \modelName{}. We only show some major results here and for more statistics please refer to Appendix \ref{sec:appendix}.

\subsection{Dataset statistics}

\begin{table}[h]
    \begin{center}
    \begin{tabular}{ |c|c|c|c| } 
     \hline
     Type & \#website & \#webpage & \#QA\\
     \hline
     KV &34 & 3207 & 168606\\ 
     Comparison & 15 & 1339 & 68578\\ 
     Table & 21 & 1901 & 163314\\
     \hline
    \end{tabular}
    \caption{Statistics of different types of websites.}
    \label{type_data_table}
    \end{center}
\end{table}

% \begin{figure*}[h]
% \centering
% \includegraphics[width=0.8\textwidth]{figs/question_dist.png} % Reduce the figure size so that it is slightly narrower than the column. Don't use precise values for figure width.This setup will avoid overfull boxes.
% \caption{The distribution of 10 most frequent trigram prefixes of questions. Figure (a) is the distribution for wh-questions without the prefix ``what is the''. Figure (b) is the distribution for yes-no questions.}
% \label{question_dis}
% \end{figure*}

The statistics of different types of websites are shown in Table \ref{type_data_table}. The most common type of website is type KV, which accounts for about a half. The least type of website is type comparison with only $17\%$ of the total websites. For we can generate questions for each value in a table, the proportion of QA pairs of type table is much bigger than its proportion of websites, which is about $40\%$. 

% \subsection{Questions in \modelName}
% \modelName{} consists of two kinds of questions: wh-questions and yes-no questions. We analyze the frequency of the leading trigram of two question types respectively. Questions starting with ``what'' are the most common questions, and questions starting with ``what is the'' account for $26.5\%$ of the whole dataset. As for yes-no questions, words like \textit{Is}, \textit{Can} and \textit{Does} are strong indicators. The length of the question is sensitive to the item name on the web page, and the average length is 8.37.

% \subsection{Answers in \modelName}

% Answers in \modelName{} are relatively short, $86.38\%$ of which are within 3 words and $55.43\%$ answers have only one word. All answers are extracted from the HTML code except for yes-no questions. However, a text that is visually a whole may be scattered in multiple HTML tags. The example shown in Figure \ref{fig-example} illustrates this phenomenon.
% The line ``\textit{21 / 26 mpg}'' is separated by ``\textit{<span>}'' tags. 
% Besides, even though an answer could be fully contained in a tag, the tag may contain additional texts. For example, the answer to the second question in Figure \ref{fig-example} is a sub-span of whole tag text \textit{2L 184hp@4800 In-Line 4}. About $3.24\%$ answers are distributed in multiple tags and $13.71\%$ answers are sub-spans of the text of HTML nodes.

\subsection{QAs in \modelName}
\modelName{} consists of two kinds of questions: wh-questions and yes-no questions. Questions starting with ``what'' are the most common questions, and questions starting with ``what is the'' account for $29.3\%$ of the whole dataset. As for yes-no questions, words like \textit{Is}, \textit{Can} and \textit{Does} are strong indicators. The average length of questions is 8.26.

Answers in \modelName{} are relatively short, $86.78\%$ of which are within 3 words and $55.21\%$ answers have only one word. However, a text that is visually a whole may be scattered in multiple HTML tags. The example shown in Figure \ref{fig-example} illustrates this phenomenon.
The line ``\textit{21 / 26 mpg}'' is separated by ``\textit{<span>}'' tags. 
Besides, a tag may contain additional texts except for the answer. For example, the answer to the second question in Figure \ref{fig-example} is a sub-span of whole tag text \textit{2L 184hp@4800 In-Line 4}. About $2.35\%$ answers are distributed in multiple tags and $13.21\%$ answers are sub-spans of the text of HTML nodes.

\section{Baseline Models}

% \begin{figure*}[h]
% \centering
% \includegraphics[width=0.95\textwidth]{figs/model.png} % Reduce the figure size so that it is slightly narrower than the column. Don't use precise values for figure width.This setup will avoid overfull boxes.
% \caption{The model architecture of V-PLM.}
% \label{mod:v-plm}
% \end{figure*}

% In our work, we propose three baseline model which gradually take more structural information into consideration. 
We propose three baseline models for WebSRC. They take different kinds of context into consideration. We describe these models in detail below.

\subsection{Pre-trained Language Model with Text (T-PLM)}
% The first baseline is to directly use Pre-trained Language Model (PLM), such as BERT, ELECTRA, etc, in the most straightforward way. 
In the first baseline, we convert the HTML code into non-structural pure text by simply deleting all HTML tags, and utilize Pre-trained Language Models (PLM), e.g. BERT \citep{devlin2019bert}, to predict answer spans. We regard it as an extractive QA task. We add two additional words \textit{yes} and \textit{no} to the end of context for yes-no questions prediction. Here the context $\mathcal{C}$ in Eq. (\ref{eq:task}) is the resulting plain text. The resulting plain text and the corresponding question are concatenated to form the input sequence $\mathbf{x}$. Then the probability distributions for each token to be the start token and the end token of the answer span can be obtained as follows:
\begin{gather}
    \mathbf{Z} = \text{PLM}\left(\mathbf{x}\right), \\ 
    \mathbf{p}^s, \mathbf{p}^e \propto \text{SoftMax}\left(\text{Linear}\left(\mathbf{Z}\right)\right),
\end{gather}
where $\mathbf{Z}$ is the resulting sequence representation calculated by PLM; $\mathbf{p}^s$ and $\mathbf{p}^e$ are start and end distributions. 
% $Softmax\left(\cdot \right)$ and $Linear\left(\cdot \right)$ are the softmax layer and the feed-forward linear layer. 
We use cross-entropy as our objective function.

In addition, after obtaining the predicted answer spans, we go over the HTML code again to find the tightest tag that contains the whole answer and take it as the predicted answer tag.

\subsection{Pre-trained Language Model with HTML (H-PLM)}
In the second baseline, we incorporate HTML tags into PLM. We called this baseline H-PLM. The model architecture of H-PLM is identical to T-PLM, the only difference is we use HTML documents with HTML tags as our context $\mathcal{C}$. To deal with the HTML tags, we remove all attributes, leaving the angle brackets, tag names, and the possible slashes unchanged. The resulting tag sequence looks like \textit{<div>, <img/>, </p>}, etc. We treat these HTML tags as new special tokens in the sequence and randomly initialize their embedding for training.

\subsection{Visual Information Enhanced Pre-trained Language Model (V-PLM)}

As introduced in Section \ref{sec:intro}, HTML is not enough to represent the whole web structure. 
In the third baseline, we take the visual information from web pages into consideration. We call this model V-PLM. It consists of three parts: PLM, visual information enhanced self-attention blocks, and a classification layer.

For each tag in HTML, we can use the bounding box provided in meta data to locate the tag in screenshot and obtain the visual embedding using the Faster R-CNN \citep{ren2016faster}.
We concatenate output hidden state $\mathbf{Z}$ from H-PLM with the corresponding visual embeddings, where tokens within the same tag share the same visual embedding. For the example in Figure \ref{fig-example}, the visual embeddings of \textit{<span>, Fuel, Economy, </span>} are all the same. For other special tokens and tokens in the question, their visual embeddings are zero vectors. 

The concatenated embedding is then fed into a self-attention block \citep{vaswani2017attention}, which is repeated $N$ times. We repeat the concatenation procedure between each self-attention block. The final representation is then sent to the classification layer to produce the starting and ending probability distribution, which is the same as H-PLM.

\section{Experiments}

\subsection{Dataset Splits}

We manually divide our dataset into train/dev/test sets at the website level, where the training set contains 50 websites, dev and test contain 10 websites respectively. Both the dev set and the test set have all three types of websites and have a similar distribution of website types. The detailed statistics of each set are shown in Table \ref{split_table}.

\begin{table}[h]
    \begin{center}
    \begin{tabular}{ |c|c|c|c| } 
     \hline
     Split & \#website & \#webpage & \#QA\\
     \hline
     Train &50 & 4549 & 307315\\ 
     Dev & 10 & 913 & 52826\\ 
     Test & 10 & 985 & 40357\\
     \hline
    \end{tabular}
    \caption{Statistics of dataset splits.}
    \label{split_table}
    \end{center}
\end{table}

\subsection{Evaluate Metrics}\label{matric}
We use three kinds of metrics for evaluation.

\begin{table*}[h]%[!htbp]
\centering
\begin{small}
    \centering
    \begin{tabular}{|l|c|c|c|c|c|c|c|c|c|}
    \hline
    \multirow{2}*{Models} & \multirow{2}*{w/ text} & \multirow{2}*{w/ tag} & \multirow{2}*{w/ screenshot} & \multicolumn{3}{c}{DEV} & \multicolumn{3}{|c|}{TEST} \\
    \cline{5-10} & & & & EM & F1 & POS & EM & F1 & POS \\
    \hline
    T-PLM (BERT) & $\surd$ &  &  &  52.12 & 61.57 & 79.74 & 39.28 & 49.49 & 67.68 \\
    H-PLM (BERT) & $\surd$ & $\surd$ &  &  61.51 & \textbf{67.04} & 82.97 & 52.61 & 59.88 & 76.13 \\
    V-PLM (BERT) & $\surd$ & $\surd$ & $\surd$ &  \textbf{62.07} & 66.66 & \textbf{83.64} & \textbf{52.84} & \textbf{60.80} & \textbf{76.39} \\
    \hline
    T-PLM (ELECTRA) & $\surd$ &  &  &  61.67 & 69.85 & 84.15 & 56.32 & 72.35 & 79.18 \\
    H-PLM (ELECTRA) & $\surd$ & $\surd$ &  &  70.12 & 74.14 & 86.33 & 66.29 & 72.21 & 83.17 \\
    V-PLM (ELECTRA) & $\surd$ & $\surd$ & $\surd$ &  \textbf{73.22} & \textbf{76.16} & \textbf{87.06} & \textbf{68.07} & \textbf{75.25} & \textbf{84.96} \\
    \hline
    \end{tabular}
\end{small}
    \caption{Experimental results of various baselines on dev and test sets. EM stands for exact match score, and POS stands for path overlap score.}
    \label{result}
\end{table*}

\textbf{Exact match (EM)}
This metric is used to evaluate whether a predicted answer is completely the same as the ground truth. It will be challenging for those answers that are only part of the tag text.

\textbf{F1 score (F1)}
This metric measures the overlap of the predicted answer and the ground truth. We split the answer and ground truth into tokens and compute the F1 score on them.

\textbf{Path overlap score (POS)}
When the model predicts an answer from a wrong tag but the text of the answer is identical to the ground truth, the exact match and F1 score will fail. Therefore we introduce path overlap score, a tag level metric that evaluates the accuracy in structure. An HTML document is a DOM tree, so for every tag, there exists a unique path from the root \textit{<HTML>} element to the tag. We compute the path overlap score (POS) between path $p_1$ and $p_2$ as following:
$\text{POS}=\frac{|P_1 \cap P_2|}{|P_1 \cup P_2|},$
where $P_1$ and $P_2$ are the sets of elements in the path $p_1$ and $p_2$ respectively. $|\cdot|$ denotes the size of a set.

\subsection{Experiment Setup}

We train our baselines on the training set and select the best models on the dev set based on the exact match score. We use uncased BERT-Base and ELECTRA-Large \citep{clark2020electra} as our backbone PLM models. The learning rate is 1e-5. The batch size is 32. We use Adam optimizer with a linear scheduler. For V-PLM, the number of self-attention blocks is 3.

% \begin{table}[h]
%     \begin{center}
%     \begin{tabular}{ |c||c|c|c| } 
%      \hline
%      Model & EM & F1 & POS\\
%      \hline
%      Text-PM & 43.91 & 55.82 & 73.74\\ 
%      Tag-PM & 49.16 & 57.7 & 70.94\\ 
%      Struc-PM & 51.17 & 59.12 & 78.58\\
%      \methodfour & \textbf{52.91} & \textbf{59.45} & \textbf{79.33}\\
%      \hline
%     \end{tabular}
%     \caption{Experiment results of baselines. EM stands for exact match score, and POS stands for path overlap score.}
%     \label{result}
%     \end{center}
% \end{table}

\subsection{Results \& Discussion}

\begin{figure}[h]
\centering
\includegraphics[width=0.48\textwidth]{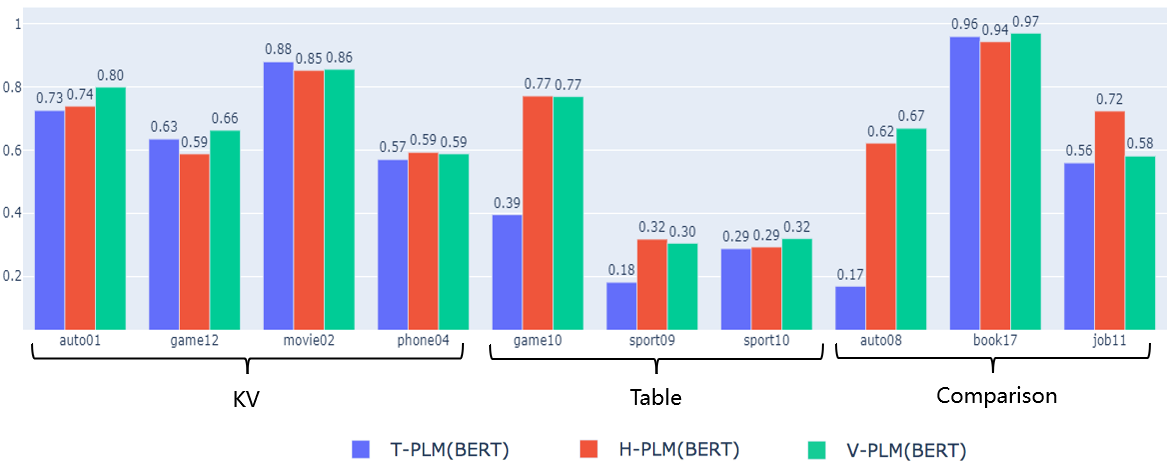} % Reduce the figure size so that it is slightly narrower than the column. Don't use precise values for figure width.This setup will avoid overfull boxes.
\caption{The performance (EM score) comparison of three baselines on 10 different websites on dev set. These websites fall into three categories: \textit{KV}, \textit{Table}, and \textit{Comparison}.}
\label{fig:performance}
\end{figure}

\begin{figure*}[h]
\centering
\includegraphics[width=1.8\columnwidth]{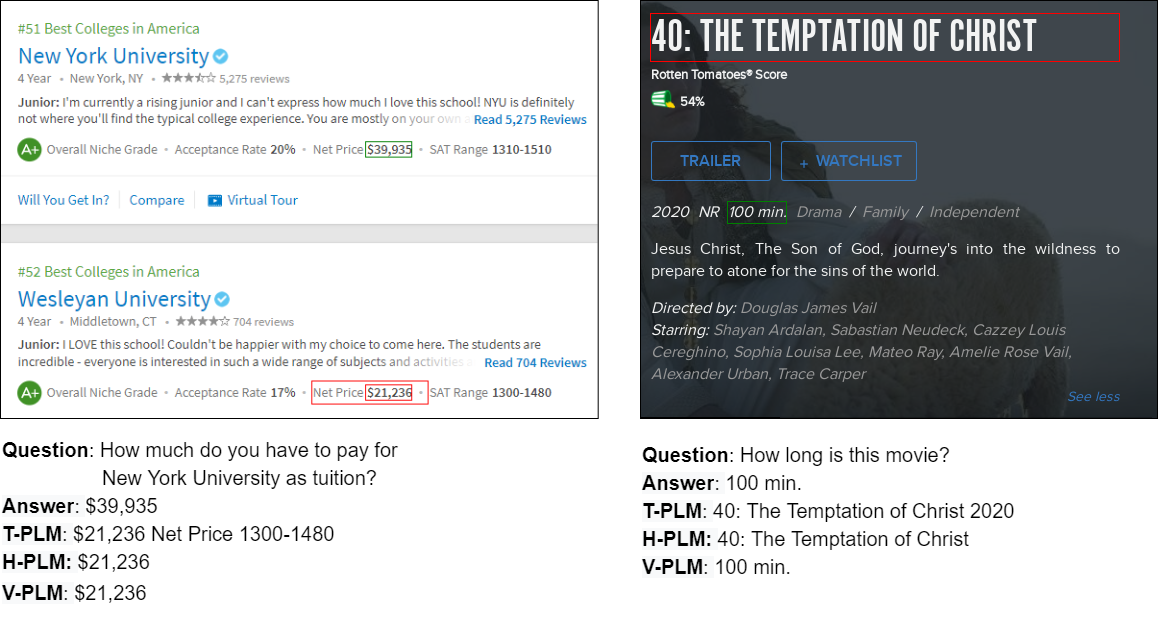} % Reduce the figure size so that it is slightly narrower than the column. Don't use precise values for figure width.This setup will avoid overfull boxes.
\caption{Case Study. The true answers are marked by green boxes, and the answers predicted by the model are marked by red boxes.}
\label{fig-case}
\end{figure*}

The experimental results are shown in Table \ref{result}. We can find that no matter which PLM is used, the more context information (i.e. text, HTML tag, screenshot) yields better performance. Specifically, comparing H-PLM with T-PLM, we find that H-PLM outperforms T-PLM by a large margin. The tag information in H-PLM can implicitly model the visual structure of web pages to some degree. Comparing V-PLM with H-PLM, we find that V-PLM can outperform H-PLM in almost all metrics, which means explicit visual features can provide additional structural information. However, we can also find that the improvement of performance is not very large. This is because the Faster-RCNN toolkit used here is pre-trained on nature images. It may not well apply to screenshots of web pages.
In the future, there is a lot of room for exploration of how to make good use of visual information. 

From Table \ref{result}, we can also find that ELECTRA-based models consistently outperform BERT-based models. ELECTRA is the best single pre-trained model on text-based MRC tasks, e.g. SQuAD2.0 \citep{rajpurkar2018know}. However, ELECTRA can achieve about 80 EM score on SQuAD2.0, while it can only achieve about 60 EM score on WebSRC. It indicates that WebSRC is still challenging for the current pre-trained language models.

In Figure \ref{fig:performance}, we further compare the performance of three baseline models on different websites.
We find that on three websites, i.e. \textit{game10}, \textit{sport09}, and \textit{auto08}, H-PLM and V-PLM outperform T-PLM with a large margin. Both \textit{game10} and \textit{sport09} fall into the category of \textit{table}, and \textit{auto08} falls into the category of \textit{comparison}. We consider in \textit{comparison} and \textit{table} websites, plain text is not enough to answer questions and more structural information is needed. Generally, to gain good performance on \textit{table} and \textit{comparison} web pages, models should have a good understanding of the global structure of web pages as well as the semantic of contents.
From Figure \ref{fig:performance}, we can also find that among three types of web pages, these models perform worst on \textit{table}.

\begin{table}[h]
    \begin{center}
    \begin{tabular}{ |c|c|c|c| } 
     \hline
     Method/Metric & \#EM & \#F1 & \#POS\\
     \hline
     T-PLM-SQuAD &29.68 & 42.91 & 62.75\\ 
     T-PLM & 54.55 & 65.28 & 76.44\\ 
     H-PLM & 61.83 & 68.24 & 78.38\\
     V-PLM & \textbf{64.71} & \textbf{69.26} & \textbf{82.81}\\
     \hline
    \end{tabular}
    \caption{Results for SQuAD model.}
    \label{squad-resut}
    \end{center}
\end{table}

Though the result has shown the difficulty of our dataset, we also wonder about the performance on our dataset of the model that does well on large-scale QA datasets. We fine-tuned a pre-trained BERT-Base model on SQuAD 2.0 dataset, and then use the parameters to initialize our baselines. For H-PLM and V-PLM we still randomly initialize the tag embedding and visual information enhanced self-attention blocks. The SQuAD model we used can achieve an exact match score of 71 in SQuAD. The result is shown in table \ref{squad-resut}. In the first row, we report the result of the SQuAD model on our dataset without fine-tuning. The exact match score is only 29.68, which means the texts from HTML are highly different from the normal textual passages. Though the fine-tuned models almost outperform the version without pretraining in all metrics, there is still a large gap between their performance in the textual QA dataset, which means we need more advanced technology to model the HTML structure.

\subsection{Case Study}

To further analyze the behaviors of our models, we select two images from our dataset and list the predictions made by baselines with a BERT backbone. The result is shown in Figure \ref{fig-case}. 
The image on the left shows information about two universities. The question asked the tuition of the first university but none of the three baselines made the right prediction. T-PLM predicted a longer string other than a raw price because there is no clear boundary of contents in the plain text, while in HTML the tags are natural separators for contents. H-PLM and V-PLM successfully fetched the entire field of \textit{Net Price}, but they failed to model the correspondence between attributes and schools and chose the wrong tuition.
The right screenshot is from a movie website, and the question is about the length of the movie. There is no leading text indicating which part would be the length, so the models need to infer the answer from structural information. Both T-PLM and H-PLM predicted the name of the movie, which means they failed to recognize the time information from plain text or HTML. V-PLM can leverage the visual hints and located the right answer. These two examples show that in order to make a comprehensive understanding of web page, a model should be able to understand the visual layout, and group the information correctly according to the spatial structure.

\section{Conclusion}
In this paper, we introduce \modelName, a multi-modal dataset for web-based structural reading comprehension with both HTML documents and screenshots. The task is to answer questions about the web pages. We evaluate several baselines on our dataset, and the results showed that incorporating layout features with textual contents is crucial to web understanding, but how to utilize such structural information requires further investigation. We hope this work can push the research on web-based structural reading comprehension forward. In the future, we will go beyond web pages to explore more structural reading comprehension tasks.

\section*{Acknowledgments}
We sincerely thank the anonymous reviewers for their valuable comments. This work has been supported by the China NSFC Projects (No. 62120106006 and No. 62106142), CCF-Tencent Open Fund and Startup Fund for Youngman Research at SJTU (SFYR at SJTU).

\bibliography{anthology,custom}

\begin{thebibliography}{47}
\expandafter\ifx\csname natexlab\endcsname\relax\def\natexlab#1{#1}\fi

\bibitem[{Antol et~al.(2015)Antol, Agrawal, Lu, Mitchell, Batra, Zitnick, and
  Parikh}]{VQA}
Stanislaw Antol, Aishwarya Agrawal, Jiasen Lu, Margaret Mitchell, Dhruv Batra,
  C.~Lawrence Zitnick, and Devi Parikh. 2015.
\newblock {VQA}: {V}isual {Q}uestion {A}nswering.
\newblock In \emph{International Conference on Computer Vision (ICCV)}.

\bibitem[{Binmakhashen and Mahmoud(2019)}]{binmakhashen2019document}
Galal~M Binmakhashen and Sabri~A Mahmoud. 2019.
\newblock Document layout analysis: A comprehensive survey.
\newblock \emph{ACM Computing Surveys (CSUR)}, 52(6):1--36.

\bibitem[{Bronzi et~al.(2013)Bronzi, Crescenzi, Merialdo, and
  Papotti}]{bronzi2013extraction}
Mirko Bronzi, Valter Crescenzi, Paolo Merialdo, and Paolo Papotti. 2013.
\newblock Extraction and integration of partially overlapping web sources.
\newblock \emph{Proceedings of the VLDB Endowment}, 6(10):805--816.

\bibitem[{Buneman(1997)}]{buneman1997semistructured}
Peter Buneman. 1997.
\newblock Semistructured data.
\newblock In \emph{Proceedings of the sixteenth ACM SIGACT-SIGMOD-SIGART
  symposium on Principles of database systems}, pages 117--121.

\bibitem[{Chakraborty et~al.(2019)Chakraborty, Lukovnikov, Maheshwari, Trivedi,
  Lehmann, and Fischer}]{chakraborty2019introduction}
Nilesh Chakraborty, Denis Lukovnikov, Gaurav Maheshwari, Priyansh Trivedi, Jens
  Lehmann, and Asja Fischer. 2019.
\newblock Introduction to neural network based approaches for question
  answering over knowledge graphs.
\newblock \emph{arXiv preprint arXiv:1907.09361}.

\bibitem[{Chang et~al.(2006)Chang, Kayed, Girgis, and
  Shaalan}]{chang2006survey}
Chia-Hui Chang, Mohammed Kayed, Moheb~R Girgis, and Khaled~F Shaalan. 2006.
\newblock A survey of web information extraction systems.
\newblock \emph{IEEE transactions on knowledge and data engineering},
  18(10):1411--1428.

\bibitem[{Chen(2018)}]{chen2018neural}
Danqi Chen. 2018.
\newblock \emph{Neural reading comprehension and beyond}.
\newblock Stanford University.

\bibitem[{Chen et~al.(2021)Chen, Lamoreaux, Wang, Durrett, Bastani, and
  Dillig}]{10.1145/3453483.3454047}
Qiaochu Chen, Aaron Lamoreaux, Xinyu Wang, Greg Durrett, Osbert Bastani, and
  Isil Dillig. 2021.
\newblock \href {https://doi.org/10.1145/3453483.3454047} {\emph{Web Question
  Answering with Neurosymbolic Program Synthesis}}, page 328–343. Association
  for Computing Machinery, New York, NY, USA.

\bibitem[{Choi et~al.(2018)Choi, He, Iyyer, Yatskar, Yih, Choi, Liang, and
  Zettlemoyer}]{choi2018quac}
Eunsol Choi, He~He, Mohit Iyyer, Mark Yatskar, Wen-tau Yih, Yejin Choi, Percy
  Liang, and Luke Zettlemoyer. 2018.
\newblock Quac: Question answering in context.
\newblock In \emph{Proceedings of the 2018 Conference on Empirical Methods in
  Natural Language Processing}, pages 2174--2184.

\bibitem[{Clark et~al.(2020)Clark, Luong, Le, and Manning}]{clark2020electra}
Kevin Clark, Minh-Thang Luong, Quoc~V. Le, and Christopher~D. Manning. 2020.
\newblock \href {https://openreview.net/pdf?id=r1xMH1BtvB} {{ELECTRA}:
  Pre-training text encoders as discriminators rather than generators}.
\newblock In \emph{ICLR}.

\bibitem[{Devlin et~al.(2019)Devlin, Chang, Lee, and
  Toutanova}]{devlin2019bert}
Jacob Devlin, Ming-Wei Chang, Kenton Lee, and Kristina Toutanova. 2019.
\newblock Bert: Pre-training of deep bidirectional transformers for language
  understanding.
\newblock In \emph{Proceedings of the 2019 Conference of the North American
  Chapter of the Association for Computational Linguistics: Human Language
  Technologies, Volume 1 (Long and Short Papers)}, pages 4171--4186.

\bibitem[{Dhingra et~al.(2017)Dhingra, Mazaitis, and Cohen}]{dhingra2017quasar}
Bhuwan Dhingra, Kathryn Mazaitis, and William~W Cohen. 2017.
\newblock Quasar: Datasets for question answering by search and reading.
\newblock \emph{arXiv preprint arXiv:1707.03904}.

\bibitem[{Dunn et~al.(2017)Dunn, Sagun, Higgins, Guney, Cirik, and
  Cho}]{dunn2017searchqa}
Matthew Dunn, Levent Sagun, Mike Higgins, V~Ugur Guney, Volkan Cirik, and
  Kyunghyun Cho. 2017.
\newblock Searchqa: A new q\&a dataset augmented with context from a search
  engine.
\newblock \emph{arXiv preprint arXiv:1704.05179}.

\bibitem[{Flesca et~al.(2004)Flesca, Manco, Masciari, Rende, and
  Tagarelli}]{flesca2004web}
Sergio Flesca, Giuseppe Manco, Elio Masciari, Eugenio Rende, and Andrea
  Tagarelli. 2004.
\newblock Web wrapper induction: a brief survey.
\newblock \emph{AI communications}, 17(2):57--61.

\bibitem[{Hao et~al.(2011)Hao, Cai, Pang, and Zhang}]{hao2011one}
Qiang Hao, Rui Cai, Yanwei Pang, and Lei Zhang. 2011.
\newblock From one tree to a forest: a unified solution for structured web data
  extraction.
\newblock In \emph{Proceedings of the 34th international ACM SIGIR conference
  on Research and development in Information Retrieval}, pages 775--784.

\bibitem[{Harley et~al.(2015)Harley, Ufkes, and
  Derpanis}]{AdamWHarly2015ICDAR_RVL_CDIP}
Adam~W. Harley, Alex Ufkes, and Konstantinos~G. Derpanis. 2015.
\newblock \href {https://doi.org/10.1109/ICDAR.2015.7333910} {Evaluation of
  deep convolutional nets for document image classification and retrieval}.
\newblock In \emph{13th International Conference on Document Analysis and
  Recognition, {ICDAR} 2015, Nancy, France, August 23-26, 2015}, pages
  991--995. {IEEE} Computer Society.

\bibitem[{Huang et~al.(2019)Huang, Chen, He, Bai, Karatzas, Lu, and
  Jawahar}]{ZhengHuang2019ICDAR_SROIE}
Zheng Huang, Kai Chen, Jianhua He, Xiang Bai, Dimosthenis Karatzas, Shijian Lu,
  and C.~V. Jawahar. 2019.
\newblock \href {https://doi.org/10.1109/ICDAR.2019.00244} {{ICDAR2019}
  competition on scanned receipt {OCR} and information extraction}.
\newblock In \emph{2019 International Conference on Document Analysis and
  Recognition, {ICDAR} 2019, Sydney, Australia, September 20-25, 2019}, pages
  1516--1520. {IEEE}.

\bibitem[{Iyyer et~al.(2017)Iyyer, Manjunatha, Guha, Vyas, Boyd-Graber, Daume,
  and Davis}]{iyyer2017amazing}
Mohit Iyyer, Varun Manjunatha, Anupam Guha, Yogarshi Vyas, Jordan Boyd-Graber,
  Hal Daume, and Larry~S Davis. 2017.
\newblock The amazing mysteries of the gutter: Drawing inferences between
  panels in comic book narratives.
\newblock In \emph{Proceedings of the IEEE Conference on Computer Vision and
  Pattern Recognition}, pages 7186--7195.

\bibitem[{Jaume et~al.(2019)Jaume, Ekenel, and
  Thiran}]{GuillaumeJaume2019ICDARW_FUNSD}
Guillaume Jaume, Hazim~Kemal Ekenel, and Jean{-}Philippe Thiran. 2019.
\newblock \href {https://doi.org/10.1109/ICDARW.2019.10029} {{FUNSD:} {A}
  dataset for form understanding in noisy scanned documents}.
\newblock In \emph{2nd International Workshop on Open Services and Tools for
  Document Analysis, OST@ICDAR 2019, Sydney, Australia, September 22-25, 2019},
  pages 1--6. {IEEE}.

\bibitem[{Joshi et~al.(2017)Joshi, Choi, Weld, and
  Zettlemoyer}]{joshi-etal-2017-triviaqa}
Mandar Joshi, Eunsol Choi, Daniel Weld, and Luke Zettlemoyer. 2017.
\newblock \href {https://doi.org/10.18653/v1/P17-1147} {{T}rivia{QA}: A large
  scale distantly supervised challenge dataset for reading comprehension}.
\newblock In \emph{Proceedings of the 55th Annual Meeting of the Association
  for Computational Linguistics (Volume 1: Long Papers)}, pages 1601--1611,
  Vancouver, Canada. Association for Computational Linguistics.

\bibitem[{Kembhavi et~al.(2017)Kembhavi, Seo, Schwenk, Choi, Farhadi, and
  Hajishirzi}]{kembhavi2017you}
Aniruddha Kembhavi, Minjoon Seo, Dustin Schwenk, Jonghyun Choi, Ali Farhadi,
  and Hannaneh Hajishirzi. 2017.
\newblock Are you smarter than a sixth grader? textbook question answering for
  multimodal machine comprehension.
\newblock In \emph{Proceedings of the IEEE Conference on Computer Vision and
  Pattern Recognition}, pages 4999--5007.

\bibitem[{Kushmerick(2000)}]{kushmerick2000wrapper}
Nicholas Kushmerick. 2000.
\newblock Wrapper induction: Efficiency and expressiveness.
\newblock \emph{Artificial intelligence}, 118(1-2):15--68.

\bibitem[{Kushmerick et~al.(1997)Kushmerick, Weld, and
  Doorenbos}]{kushmerick1997wrapper}
Nicholas Kushmerick, Daniel~S Weld, and Robert Doorenbos. 1997.
\newblock \emph{Wrapper induction for information extraction}.
\newblock University of Washington Washington.

\bibitem[{Lai et~al.(2017)Lai, Xie, Liu, Yang, and Hovy}]{lai2017race}
Guokun Lai, Qizhe Xie, Hanxiao Liu, Yiming Yang, and Eduard Hovy. 2017.
\newblock Race: Large-scale reading comprehension dataset from examinations.
\newblock In \emph{Proceedings of the 2017 Conference on Empirical Methods in
  Natural Language Processing}, pages 785--794.

\bibitem[{Lewis et~al.(2006)Lewis, Agam, Argamon, Frieder, Grossman, and
  Heard}]{DavidDLewis2006SIGIR_IIT_CDIP}
David~D. Lewis, Gady Agam, Shlomo Argamon, Ophir Frieder, David~A. Grossman,
  and Jefferson Heard. 2006.
\newblock \href {https://doi.org/10.1145/1148170.1148307} {Building a test
  collection for complex document information processing}.
\newblock In \emph{{SIGIR} 2006: Proceedings of the 29th Annual International
  {ACM} {SIGIR} Conference on Research and Development in Information
  Retrieval, Seattle, Washington, USA, August 6-11, 2006}, pages 665--666.
  {ACM}.

\bibitem[{Li et~al.(2020)Li, Xu, Cui, Huang, Wei, Li, and Zhou}]{li2020docbank}
Minghao Li, Yiheng Xu, Lei Cui, Shaohan Huang, Furu Wei, Zhoujun Li, and Ming
  Zhou. 2020.
\newblock Docbank: A benchmark dataset for document layout analysis.
\newblock In \emph{Proceedings of the 28th International Conference on
  Computational Linguistics}, pages 949--960.

\bibitem[{Li et~al.(2016)Li, Li, He, Wang, Cao, Zhou, and Xu}]{li2016dataset}
Peng Li, Wei Li, Zhengyan He, Xuguang Wang, Ying Cao, Jie Zhou, and Wei Xu.
  2016.
\newblock Dataset and neural recurrent sequence labeling model for open-domain
  factoid question answering.
\newblock \emph{arXiv preprint arXiv:1607.06275}.

\bibitem[{Lockard et~al.(2019)Lockard, Shiralkar, and
  Dong}]{lockard2019openceres}
Colin Lockard, Prashant Shiralkar, and Xin~Luna Dong. 2019.
\newblock Openceres: When open information extraction meets the semi-structured
  web.
\newblock In \emph{Proceedings of the 2019 Conference of the North American
  Chapter of the Association for Computational Linguistics: Human Language
  Technologies, Volume 1 (Long and Short Papers)}, pages 3047--3056.

\bibitem[{Mathew et~al.(2021)Mathew, Karatzas, and Jawahar}]{docvqa_wacv}
Minesh Mathew, Dimosthenis Karatzas, and C.V. Jawahar. 2021.
\newblock Docvqa: A dataset for vqa on document images.
\newblock In \emph{WACV}, pages 2200--2209.

\bibitem[{Mishra et~al.(2019)Mishra, Shekhar, Singh, and
  Chakraborty}]{mishraICDAR19}
Anand Mishra, Shashank Shekhar, Ajeet~Kumar Singh, and Anirban Chakraborty.
  2019.
\newblock Ocr-vqa: Visual question answering by reading text in images.
\newblock In \emph{ICDAR}.

\bibitem[{Muslea et~al.(1999)Muslea, Minton, and
  Knoblock}]{muslea1999hierarchical}
Ion Muslea, Steve Minton, and Craig Knoblock. 1999.
\newblock A hierarchical approach to wrapper induction.
\newblock In \emph{Proceedings of the third annual conference on Autonomous
  Agents}, pages 190--197.

\bibitem[{Rajpurkar et~al.(2018)Rajpurkar, Jia, and Liang}]{rajpurkar2018know}
Pranav Rajpurkar, Robin Jia, and Percy Liang. 2018.
\newblock Know what you don’t know: Unanswerable questions for squad.
\newblock In \emph{Proceedings of the 56th Annual Meeting of the Association
  for Computational Linguistics (Volume 2: Short Papers)}, pages 784--789.

\bibitem[{Rajpurkar et~al.(2016)Rajpurkar, Zhang, Lopyrev, and
  Liang}]{rajpurkar2016squad}
Pranav Rajpurkar, Jian Zhang, Konstantin Lopyrev, and Percy Liang. 2016.
\newblock Squad: 100,000+ questions for machine comprehension of text.
\newblock In \emph{Proceedings of the 2016 Conference on Empirical Methods in
  Natural Language Processing}, pages 2383--2392.

\bibitem[{Reddy et~al.(2019)Reddy, Chen, and Manning}]{reddy2019coqa}
Siva Reddy, Danqi Chen, and Christopher~D Manning. 2019.
\newblock Coqa: A conversational question answering challenge.
\newblock \emph{Transactions of the Association for Computational Linguistics},
  7:249--266.

\bibitem[{Ren et~al.(2016)Ren, He, Girshick, and Sun}]{ren2016faster}
Shaoqing Ren, Kaiming He, Ross Girshick, and Jian Sun. 2016.
\newblock Faster r-cnn: Towards real-time object detection with region proposal
  networks.
\newblock \emph{IEEE transactions on pattern analysis and machine
  intelligence}, 39(6):1137--1149.

\bibitem[{Singh et~al.(2019)Singh, Natarajan, Shah, Jiang, Chen, Batra, Parikh,
  and Rohrbach}]{singh2019towards}
Amanpreet Singh, Vivek Natarajan, Meet Shah, Yu~Jiang, Xinlei Chen, Dhruv
  Batra, Devi Parikh, and Marcus Rohrbach. 2019.
\newblock Towards vqa models that can read.
\newblock In \emph{Proceedings of the IEEE Conference on Computer Vision and
  Pattern Recognition}, pages 8317--8326.

\bibitem[{Talmor et~al.(2019)Talmor, Herzig, Lourie, and
  Berant}]{talmor2019commonsenseqa}
Alon Talmor, Jonathan Herzig, Nicholas Lourie, and Jonathan Berant. 2019.
\newblock Commonsenseqa: A question answering challenge targeting commonsense
  knowledge.
\newblock In \emph{Proceedings of the 2019 Conference of the North American
  Chapter of the Association for Computational Linguistics: Human Language
  Technologies, Volume 1 (Long and Short Papers)}, pages 4149--4158.

\bibitem[{Tapaswi et~al.(2016)Tapaswi, Zhu, Stiefelhagen, Torralba, Urtasun,
  and Fidler}]{tapaswi2016movieqa}
Makarand Tapaswi, Yukun Zhu, Rainer Stiefelhagen, Antonio Torralba, Raquel
  Urtasun, and Sanja Fidler. 2016.
\newblock Movieqa: Understanding stories in movies through question-answering.
\newblock In \emph{Proceedings of the IEEE conference on computer vision and
  pattern recognition}, pages 4631--4640.

\bibitem[{Vaswani et~al.(2017)Vaswani, Shazeer, Parmar, Uszkoreit, Jones,
  Gomez, Kaiser, and Polosukhin}]{vaswani2017attention}
Ashish Vaswani, Noam Shazeer, Niki Parmar, Jakob Uszkoreit, Llion Jones,
  Aidan~N Gomez, {\L}ukasz Kaiser, and Illia Polosukhin. 2017.
\newblock Attention is all you need.
\newblock In \emph{Advances in neural information processing systems}, pages
  5998--6008.

\bibitem[{Yagcioglu et~al.(2018)Yagcioglu, Erdem, Erdem, and
  Ikizler-Cinbis}]{yagcioglu2018recipeqa}
Semih Yagcioglu, Aykut Erdem, Erkut Erdem, and Nazli Ikizler-Cinbis. 2018.
\newblock Recipeqa: A challenge dataset for multimodal comprehension of cooking
  recipes.
\newblock In \emph{Proceedings of the 2018 Conference on Empirical Methods in
  Natural Language Processing}, pages 1358--1368.

\bibitem[{Yang et~al.(2018)Yang, Qi, Zhang, Bengio, Cohen, Salakhutdinov, and
  Manning}]{yang2018hotpotqa}
Zhilin Yang, Peng Qi, Saizheng Zhang, Yoshua Bengio, William Cohen, Ruslan
  Salakhutdinov, and Christopher~D Manning. 2018.
\newblock Hotpotqa: A dataset for diverse, explainable multi-hop question
  answering.
\newblock In \emph{Proceedings of the 2018 Conference on Empirical Methods in
  Natural Language Processing}, pages 2369--2380.

\bibitem[{Zanibbi et~al.(2004)Zanibbi, Blostein, and Cordy}]{zanibbi2004survey}
Richard Zanibbi, Dorothea Blostein, and James~R Cordy. 2004.
\newblock A survey of table recognition.
\newblock \emph{Document Analysis and Recognition}, 7(1):1--16.

\bibitem[{Zeng et~al.(2020)Zeng, Li, Li, Hu, and Hu}]{zeng2020survey}
Changchang Zeng, Shaobo Li, Qin Li, Jie Hu, and Jianjun Hu. 2020.
\newblock A survey on machine reading comprehension—tasks, evaluation metrics
  and benchmark datasets.
\newblock \emph{Applied Sciences}, 10(21):7640.

\bibitem[{Zhang et~al.(2020)Zhang, Shou, Pei, Gong, Wen, and
  Jiang}]{zhang-etal-2020-graph}
Xingyao Zhang, Linjun Shou, Jian Pei, Ming Gong, Lijie Wen, and Daxin Jiang.
  2020.
\newblock \href {https://doi.org/10.18653/v1/2020.coling-main.5} {A graph
  representation of semi-structured data for web question answering}.
\newblock In \emph{Proceedings of the 28th International Conference on
  Computational Linguistics}, pages 51--61, Barcelona, Spain (Online).
  International Committee on Computational Linguistics.

\bibitem[{Zhong et~al.(2019)Zhong, Tang, and
  Jimeno{-}Yepes}]{XuZhong2019ICDAR_PubLayNet}
Xu~Zhong, Jianbin Tang, and Antonio Jimeno{-}Yepes. 2019.
\newblock \href {https://doi.org/10.1109/ICDAR.2019.00166} {Publaynet: Largest
  dataset ever for document layout analysis}.
\newblock In \emph{2019 International Conference on Document Analysis and
  Recognition, {ICDAR} 2019, Sydney, Australia, September 20-25, 2019}, pages
  1015--1022. {IEEE}.

\bibitem[{Zhu et~al.(2005)Zhu, Nie, Wen, Zhang, and Ma}]{zhu20052d}
Jun Zhu, Zaiqing Nie, Ji-Rong Wen, Bo~Zhang, and Wei-Ying Ma. 2005.
\newblock 2d conditional random fields for web information extraction.
\newblock In \emph{Proceedings of the 22nd international conference on Machine
  learning}, pages 1044--1051.

\bibitem[{Zhu et~al.(2006)Zhu, Nie, Wen, Zhang, and Ma}]{zhu2006simultaneous}
Jun Zhu, Zaiqing Nie, Ji-Rong Wen, Bo~Zhang, and Wei-Ying Ma. 2006.
\newblock Simultaneous record detection and attribute labeling in web data
  extraction.
\newblock In \emph{Proceedings of the 12th ACM SIGKDD international conference
  on Knowledge discovery and data mining}, pages 494--503.

\end{thebibliography}
\bibliographystyle{acl_natbib}

% \newpage
\clearpage
\newpage
\appendix

\section{Appendix}
\label{sec:appendix}
\subsection{Dataset distribution}
\begin{figure}[h]
\centering
\includegraphics[width=0.5\textwidth]{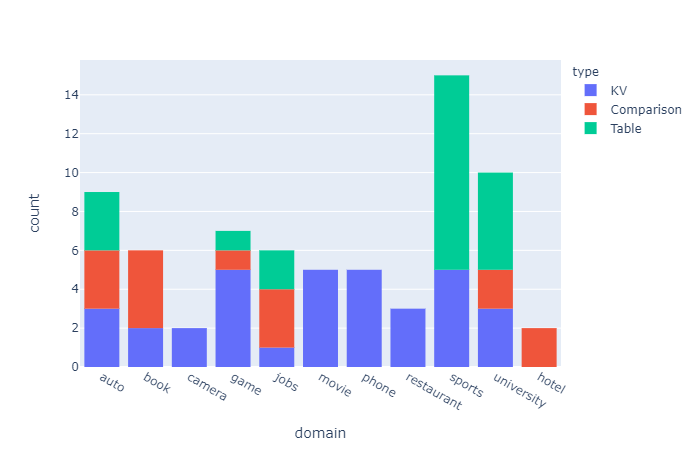} 
\caption{Distribution of different type of websites in different domains.}
\label{type_domain}
\end{figure}

The distribution of different website types in various domains is shown in Figure \ref{type_domain}. As the figure illustrated, not all domains contain three website types while type KV almost exists in all domains. Websites
of type comparison are concentrated in the domain of goods, i.e. \textit{auto}, \textit{book}, in the form of item comparison. Most web pages with type table belong to the domain \textit{sports}, which contain the score data of players.

\begin{figure}[t]
\centering
\includegraphics[width=0.9\columnwidth]{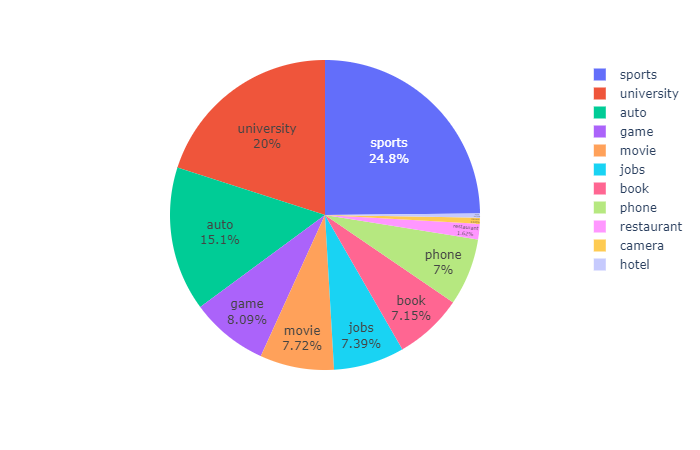}
% Reduce the figure size so that it is slightly narrower than the column. Don't use precise values for figure width.This setup will avoid overfull boxes.
\caption{Data distribution in different domains.}
\label{domain_data_table}
\end{figure}

Figure \ref{domain_data_table} shows the data distribution in different domains. Domains \textit{auto}, \textit{university} and \textit{sports} account for more than half of the data, and \textit{hotel}, \textit{camera} and \textit{restaurant} are the domains that with least data. This distribution can attribute to the amount of information carried by websites and the amount of information in different domains that are interested by people.

\subsection{HTML statistics}

\begin{figure}[t]
\centering
\includegraphics[width=1\columnwidth]{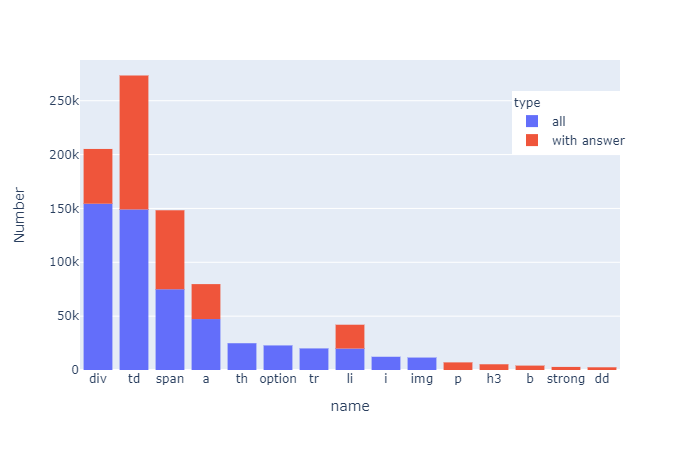}
% Reduce the figure size so that it is slightly narrower than the column. Don't use precise values for figure width.This setup will avoid overfull boxes.
\caption{Distribution of HTML tags. The blue is the top 10 HTML tags in all HTML code, and the red is the top 10 HTML tags containing an answer.} 
\label{tag-dist}
\end{figure}

We explored the distribution of HTML tags in \modelName{}. Figure \ref{tag-dist} shows the relative proportion of top 10 frequent HTML tags and top 10 frequent HTML tags containing an answer. Three most common tags are \textit{<div>}, \textit{<td>} and \textit{<span>} on all pages, which are also most frequent tags containing answers. \textit{<div>} and \textit{<span>} are used for separating an area, while \textit{<td>} represents a table cell. This observation indicates that the type of tag may imply the semantics of the content. Though \textit{<div>} is the most frequent tag, \textit{<td>} is much more likely to contain an answer, for the reason that \textit{<div>} is often used in framing the web page while \textit{<td>} is commonly used for presenting a value. The average number of HTML tags in web pages is 177. The mean depth of HTML DOM trees is 9.8 and the mean depth of tags containing answers is 7.1, which means the upper nodes in the DOM tree would provide more structural information and the lower nodes would contain more specific information.

% \subsection{Questions in \modelName}
% We visualize the distribution of the trigram prefix of ``wh-'' questions without the prefix ``what is the'' in Figure \ref{question_dis}(a). The trigram prefix distribution of yes-no questions is shown in Figure \ref{question_dis}(b). 

% \begin{figure*}[h]
% \centering
% \includegraphics[width=0.8\textwidth]{figs/question_dist.png} % Reduce the figure size so that it is slightly narrower than the column. Don't use precise values for figure width.This setup will avoid overfull boxes.
% \caption{The distribution of 10 most frequent trigram prefixes of questions. Figure (a) is the distribution for wh-questions without the prefix ``what is the''. Figure (b) is the distribution for yes-no questions.}
% \label{question_dis}
% \end{figure*}

\end{document}